\documentclass[10pt,letterpaper]{article}

\usepackage{cogsci}

\cogscifinalcopy

\usepackage{pslatex}
\usepackage{apacite}
\usepackage{graphicx}
\usepackage{linguex}
\usepackage{float}

\title{FutureVision: A methodology for the investigation of future cognition}

\author{
\begin{tabular}{c} {\bf Tiago Timponi Torrent} {(tiago.torrent@ufjf.br)} \\ FrameNet Brasil \\ Federal University of Juiz de Fora - CNPq \\ \\
{\bf Nicolás Hinrichs} {(hinrichsn@cbs.mpg.de)} \\ Research Group Cognition and Plasticity \\ Max Planck Inst. for Human Cognitive and Brain Sciences \\ \\
{\bf Igor Lourenço} {(ials@ufu.br)} \\ Institute for Language and Linguistics \\ Federal University of Uberlândia - CNPq \\ \\
{\bf Marcelo Viridiano} {(marcelo.viridiano@case.edu)} \\ FrameNet Brasil \\ Federal University of Juiz de Fora - CNPq \end{tabular}
\begin{tabular}{c} {\bf Mark Turner} {(turner@case.edu)} \\ Red Hen Lab \\ Case Western Reserve University \\ \\
{\bf Frederico Belcavello} {(fred.belcavello@ufjf.br)} \\ FrameNet Brasil \\ Federal University of Juiz de Fora - CNPq \\ \\
{\bf Arthur Lorenzi Almeida} {(arthur.lorenzi@estudante.ufjf.br)} \\ FrameNet Brasil \\ Federal University of Juiz de Fora \\ \\
{\bf Ely Edison Matos} {(ely.matos@ufjf.br)} \\ FrameNet Brasil \\ Federal University of Juiz de Fora \end{tabular} 
}
\date{}

\begin{document}

\maketitle

\begin{abstract}
This paper presents a methodology combining multimodal semantic analysis with an eye-tracking experimental protocol to investigate the cognitive effort involved in understanding the communication of future scenarios. We conduct a pilot study examining how visual fixation patterns vary during evaluation of valence and counterfactuality in fictional ad pieces describing futuristic scenarios, using a portable eye tracker. Participants' eye movements are recorded while evaluating the stimuli and describing them to a conversation partner. Gaze patterns are analyzed alongside semantic representations of the stimuli and participants' descriptions, constructed from a frame semantic annotation of both linguistic and visual modalities. Preliminary results show that far-future and pessimistic scenarios are associated with longer fixations and more erratic saccades, supporting the hypothesis that fractures in the base spaces underlying interpretation of future scenarios increase cognitive load for comprehenders.

\end{abstract}

\section{Introduction}

Speculative Futures and Quantitative Futurism \cite{webbsignals,hoffman2022speculative,mcgonigal2022imaginable} offer future scenarios that help people and organizations plan toward long-term goals. These fields devise methodologies for scenario forecasting and propose interventions—games, immersive experiences, fictional ad pieces––to foster future-oriented thinking. Such methodologies usually aim at both building comprehensive future scenarios and communicating them effectively.

\begin{figure}[t]
\begin{center}
\includegraphics[width=\columnwidth]{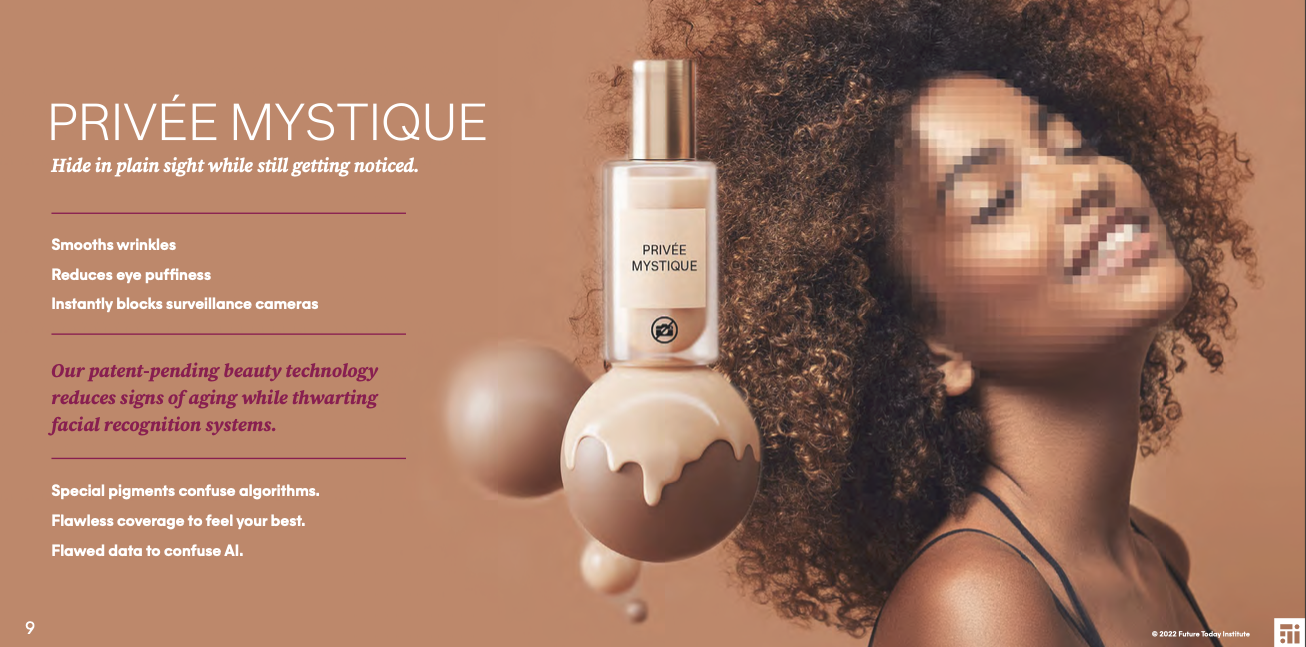}
\end{center}
\caption{Fictional ad from the 2022 Future Today Institute Tech Trends Report.} 
\label{fig:mystique}
\end{figure}

Human cognition relies on a range of processes to construct, interpret, and communicate counterfactual and imagined scenarios. This paper presents a methodology aimed at investigating factors that impact the cognitive effort required to understand multimodal communication of futuristic scenarios. We focus on evaluating how violations of semantic frames used as foundational structures for analogical projections influence cognitive effort. To this end, we conduct semantic annotation of fictional ad pieces such as the one in Figure \ref{fig:mystique} and evaluate cognitive effort required to interpret the future scenarios conveyed by such ad pieces using an eye-tracking protocol.

\citeA{fauconnier1997mappings} proposes mental spaces as representations of human on-the-fly cognition. In mental spaces theory, as humans access semiotic clues––whether linguistic, visual, or of any other type––they organize schemata for modeling and elaborating on the situation they need to make sense of. The author emphasizes the power of base spaces––the starting points of any mental space network––and the dependence of mental space networks on base frames. \citeA{Fillmore1982}, in turn, argues that human cognition relies on frames for meaning making. In Frame Semantics, frames are defined as systems of shared and interrelated concepts that structure our experience of the world. Frames permit alternative perspectives \cite{Fillmore1985} and can be organized into a network-like database \cite{Baker:1998:BFP:980845.980860,baker2003}. In this sense, a notion such as that of \textit{risk} can be modeled in terms of a scene where there is an asset that can suffer––or not––some harm, depending on how safe it is in a given situation. This scene can be further ellaborated from the perspective of the asset, which can be at \textit{risk}, or the situation, which can be \textit{risky}. As pointed out by \citeA{fillmore:_towar}, those different perspectives are evoked by the noun \textit{risk} and by the adjective \textit{risky}, respectively.

\citeA{torrent&turnerfuturemind} find that, even as mental space networks evolve, extend, and reform during counterfactual thinking, the base spaces and frames structuring them generally persist. The mental operations and communicative constructions on which they depend favor the persistence of the base. The base is especially important for imagining and understanding unreal worlds, such as counterfactual, future, or fictional scenarios. 

In this research, we propose a methodology––FutureVision––to gauge the extent to which violations of the base spaces used for futuristic projections correlate with an increase in the cognitive effort required to explain scenarios presented in the form of fictitious advertising pieces. The methodology builds upon the theoretical-methodological foundations of Cognitive Linguistics \cite{Fillmore1982,Fillmore1985,fauconnier1997mappings,fauconnier2002way}, multimodal frame-semantic annotation \cite{belcavello-etal-2020-frame,belcavello-etal-2022-charon}, and eye-tracking techniques \cite{10.5555/3134162,Conklinetal2018}. We analyze and discuss data from a pilot experiment using the methodology.

\section{Persistence of The Base in Future Cognition}

Analogical and projective processes \cite{fauconnier2002way} anchor future scenarios in experiences culturally shared by interlocutors involved in a communicative process. \citeA{torrent&turnerfuturemind} demonstrate specific patterns of basic structure persistence during scenario forecasting.

Consider the fictional ad piece from the 2022 Future Today Institute Tech Trends Report \cite{fti2022} in Figure \ref{fig:mystique}. It tells a story of the future. Its story teller has a future in mind--Future A--and wants the story the audience constructs--Future B--to be maximally similar. To make that happen, the storyteller must guide the audience to activate certain background knowledge. The storyteller establishes a scene of blended joint attention \cite{Turner2015}: the storyteller and the audience are jointly attending, not to the here and now of the conversation, but rather to a there and then in the future.  

The storyteller prompts the audience to build mental spaces needed to comprehend the ad. These mental spaces are triggered by linguistic and visual elements and are structured by frames. In this case, the frame is \texttt{Body\_decoration}. Frame elements map to elements in the ad. The \textsc{Decoration} is the Privée Mystique, the \textsc{Decorated\_individual} is the model, and her pixelated face is the \textsc{Body\_location}.\footnote{We use \texttt{Courier} font to indicate a frame. Complete descriptions of frames cited in this paper can be found at \texttt{http://webtool.frame.net.br/report/frame}}

But of course, the cosmetic foundation offered by the ad does not yet exist; it is a future product. The ad was published in the 2022 FTI Report as an example of a near-future optimistic scenario, and this communicative context prompts the audience to create a mental space for 50 years from now, when the product is bought by people represented by the model. 

The audience is invited to build other mental spaces. Some might be in the past, for the history of the evolution of make-up, given its uses through time--in ceremonial events or in theatrical performances in lieu of masks. One might be for now. Others might be 50 years from now. One of those mental spaces 50 years from now might have facial recognition technology against which there is no defense. But one of them will have Privée Mystique, which protects against facial recognition. Since the notion of concealing still persists, a generic mental space can operate over all the spaces that involve concealment.

Vital relations \cite{fauconnier2002way} connect elements in the spaces. The different people wearing make-up throughout the years are linked via \textbf{Analogy}, which allows for the compression of all the different instances of make-up users into one single \textbf{Identity}. \textbf{Disanalogy} connects a product primarily made to conceal flaws in the skin surface to one aimed at concealing identities. To be able to make this connection, the audience must bring into play yet another frame: \texttt{Protecting}, where an \textsc{Asset} is protected from some \textsc{Danger} by the deployment of some \textsc{Protection}. These new connections build an access point for the construction of a new future concept.

From that access point, the audience will be able to reconstrue the frame blending intended by the storyteller. The clues for building this new blended space in the future include linguistic elements––\textit{"Hide in plain sight while still getting noticed"}––and visual elements––the camera icon in the foundation bottle and the pixelated aspect of the model's face. Other less direct clues help the audience fit the new frame into their system of concepts: the model in the ad is black and black women are amongst the most prominent victims of harmful algorithmic biases.

\begin{figure}[t]
\begin{center}
\includegraphics[width=\columnwidth]{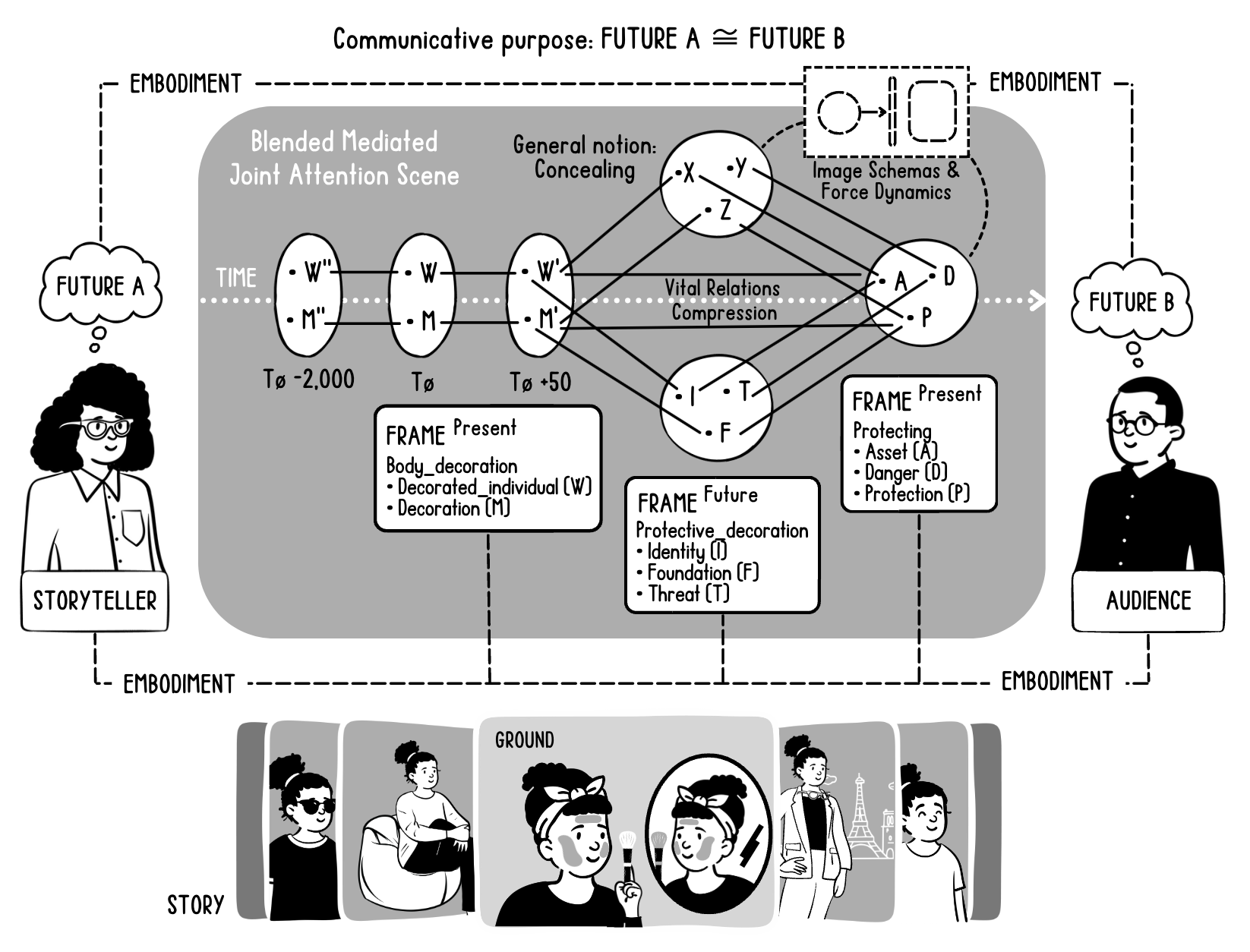}
\end{center}
\caption{Cognitive mechanisms involved in the comprehension of the fictional ad piece.} 
\label{fig:persistence}
\end{figure}

During this construction of meaning, storytellers and audience members rely on embodiment \cite{kirchhoff2018} to recruit frames and reframe concepts. They deploy image schemas and force dynamics \cite{talmy1988force} to make sense of generalities. The make-up is conceived of as a barrier stopping an agonist––the face recognition technology––from invading a bounded region––the model's identity. Figure \ref{fig:persistence} summarizes the cognitive mechanisms involved in the comprehension of the ad. 

Not every attempt of communicating some future scenario is successful. The question to be asked is: why are certain future scenarios more easily communicated and understood than others? To this end, in the next section, we present FutureVision: a methodology for investigating how the cognitive effort needed to understand communication of a future scenario depends on differences in shared background knowledge, counterfactuality, and valence.  

\section{FutureVision: A Methodology for Investigating Future Cognition}

The FutureVision methodology builds on recent advances in multimodal frame-semantic annotation \cite{belcavello-etal-2020-frame,belcavello-etal-2022-charon} and eye-tracking techniques \cite{10.5555/3134162,Conklinetal2018} to shed light on the cognitive effort involved in the comprehension of future scenarios.

To test the feasibility of the methodology, a pilot study deployed eye-tracking glasses to evaluate gaze fixation patterns during scenario evaluation.\footnote{The experimental protocol was approved by the Committee for Ethics in Research at the Federal University of Juiz de Fora under the number 84376424.0.0000.5147.} A conversation pair was formed at random, based on availability. This pair was invited to enter a room, sit in chairs about 50 centimeters apart at an approximate 90-degree angle, in front of a 40-inch video monitor equipped with an image capture camera. In this configuration, participants were in a comfortable position both to look at each other and to look at the monitor. They were also in the camera's visual field. 

Participant A, chosen at random from one of the two participants, wore the Tobii Pro Glasses 3 eye tracker glasses and assumed the role of "speaker". Participant B, without the glasses, assumed the role of "listener". Participants were asked to read instructions explaining that the experiment would last approximately 25 minutes, with an optional break of 5 minutes; that the monitor would display 8 advertisements (text and image) for products that do not exist now but that could, in principle, exist in the future.\footnote{For the pilot experiment, text in the ads was translated into Brazilian Portuguese.} The ads were selected from the Future Today Institute's 2022 Technology Trends Report \cite{fti2022} and were grouped into four sets according to Valence (pessimistic/optimistic) and Counterfactuality (near/far future).\footnote{Data and scripts are available via \texttt{https://osf.io/9kvu2}.} Participant A's task was to explain to participant B the future scenario that brings relevance to whatever product or service the ad presents. The data collected were the fixation points of participant A's gaze on the screen and the utterances participant A made–– transcribed into text.

\subsection{Participant selection}

Participants were recruited and selected based on their knowledge of Brazilian Portuguese, the language used in the experiment. Inclusion criteria required participants to be over 18 years of age, native speakers of Brazilian Portuguese, literate, and without hearing complaints or visual problems (unless the visual problems could be corrected by the use of lenses). Subjects were not filtered or separated by sex or gender. Participation in the experiment was free and consensual.

Subjects were excluded from participating if they had evidence or history of central nervous system disease, problematic alcohol consumption, abuse of other substances, learning or memory problems, psychiatric disorder, serious medical illness, or photosensitive epilepsy, or were taking  medication known to have centrally acting effects. Compliance with these requirements was measured via a triage questionnaire.




\subsection{Data analysis}

\subsubsection{Multimodal frame-semantic annotation of experimental stimuli}

Eight fictional ads from the FTI 2022 Tech Trends Report––two for each combination: near-future optimistic, near-future pessimistic, far-future optimistic and far-future pessimistic––were annotated by two experts with 10+ years of experience in annotation for frames and frame elements for both the linguistic and the visual communicative modes. Semantic representations built from the annotations were stored for each stimulus. 

\subsubsection{Frame-semantic annotation of Participant A's descriptions}

Participant A's verbal descriptions of the stimuli were automatically transcribed and annotated using the categories of frame semantics.\footnote{Participants descriptions were transcribed using Notta.ai (available at \texttt{https://app.notta.ai/}) and manually revised by a trained linguist. Frame-semantic annotation was performed on an instance of LOME \cite{xia-etal-2021-lome} trained with FrameNet Brasil data.} This involves identifying the main conceptual structures (frames) that correspond to the participants' linguistic output, providing a qualitative context for the gaze pattern data.

\subsubsection{Processing eye-tracking data}

Raw gaze tracking data was extracted from Tobii Pro Glasses 3, focusing on key metrics such as fixation duration, fixation count and saccadic movement patterns. Gaze data was segmented according to the different stimuli presented during the session, allowing a detailed comparison of fixation patterns in various valence and counterfactual conditions. Areas of Interest (AOIs) were defined so as to keep track of the multimodal composition of the ads, including both text blocks and profiled images, as well as the whole screen.

\subsubsection{Statistical analysis}

Mixed-effects regression models were employed to analyze the relationship between gaze behavior, cognitive effort, and scenario interpretation. Fixation duration, pupil dilation, and AOI transition probabilities were treated as dependent variables, while scenario valence and future distance served as fixed effects. Random intercepts accounted for individual differences in cognitive flexibility and baseline attention patterns. Mediation analysis tested whether pupil dilation mediated the relationship between scenario distance and semantic similarity scores. Cluster analysis was conducted to identify gaze behavior patterns, classifying participants into low and high cognitive load groups based on fixation variability and saccadic patterns. Markov chain models were applied to Areas of Interest (AOIs) transition data to quantify gaze regressions and assess conceptual integration difficulty across conditions. 

\subsubsection{Integration of behavioral data and gaze pattern data}

Following the methodology proposed by \citeA{viridiano-etal-2022-case}, cosine similarity between the frame-semantic representations of the stimuli and that of the participants' descriptions was calculated to assess how close their interpretations are to the intended valence and counterfactuality. This is correlated with fixation variability to explore how cognitive costs influence perception and interpretation processes. The results of the comparative and regression analyses are interpreted to understand how the fracture of the base space manifests itself in attachment patterns and cognitive processing.

\section{Results}

In this section we present the results obtained from the pilot application of the FutureVision methodology to three pairs of participants.

\subsection{Similarities between semantic representations}

Cosine similarities for the frame-semantic representations obtained from the annotation of the stimuli and of the participants' descriptions of the fictional ads are given in Table \ref{tbl:cosine}. The first trend to be noted is that near-future scenarios present, on average, higher cosine similarity scores than far-future scenarios. Also, optimistic scenarios tend to have higher similarity scores than pessimistic ones.

Also noteworthy is the discrepancy between the similarity scores for the two near-future optimistic scenarios. One of the fictional ads in this condition is precisely the Privée Mystique ad (Figure \ref{fig:mystique}). Although the FTI Tech Trends Report \cite{fti2022} presents this as an optimistic scenario, since it would allow population profiles more prone to suffering from algorithmic bias to protect themselves against harmful technologies, all three participants mentioned in their descriptions various frames other than \texttt{Protecting}, namely: \texttt{Personal\_relationship}, evoked by lexical units such as \textit{traição.n 'betrayal.n'}, and \texttt{Theft}, evoked by \textit{roubo.n 'theft.n'}, as shown in \ref{ex:betrayal}.
\ex.\label{ex:betrayal} Acho que é uma visão negativa, justamente porque me remete muito a questão da \textbf{traição ou roubo}, porque não vejo sentido de você usar uma peça de beleza para poder se esconder de câmeras. (\textit{I think it is a negative perspective [on the future], precisely because it leads me into thinking about the issues of \textbf{betrayal or theft}, because I see no reason for you to wear some beauty product to hide from cameras.})

\begin{figure}[h]
\begin{center}
\includegraphics[width=\columnwidth]{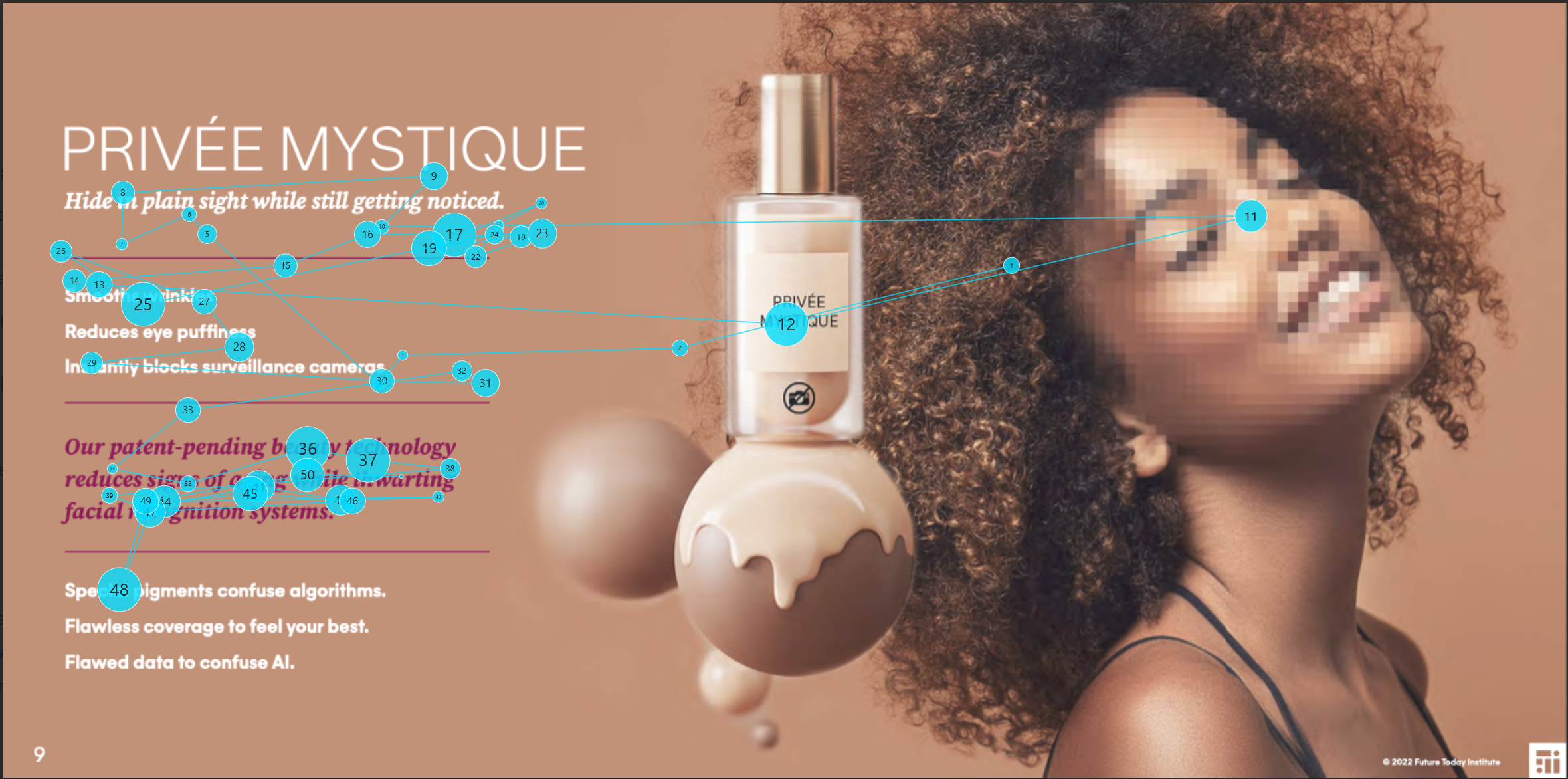}
\end{center}
\caption{PA1 gaze plot for the Privée Mystique ad for the first 15 seconds.} 
\label{fig:mystique_gazeplot}
\end{figure}

\begin{figure}[h]
\begin{center}
\includegraphics[width=\columnwidth]{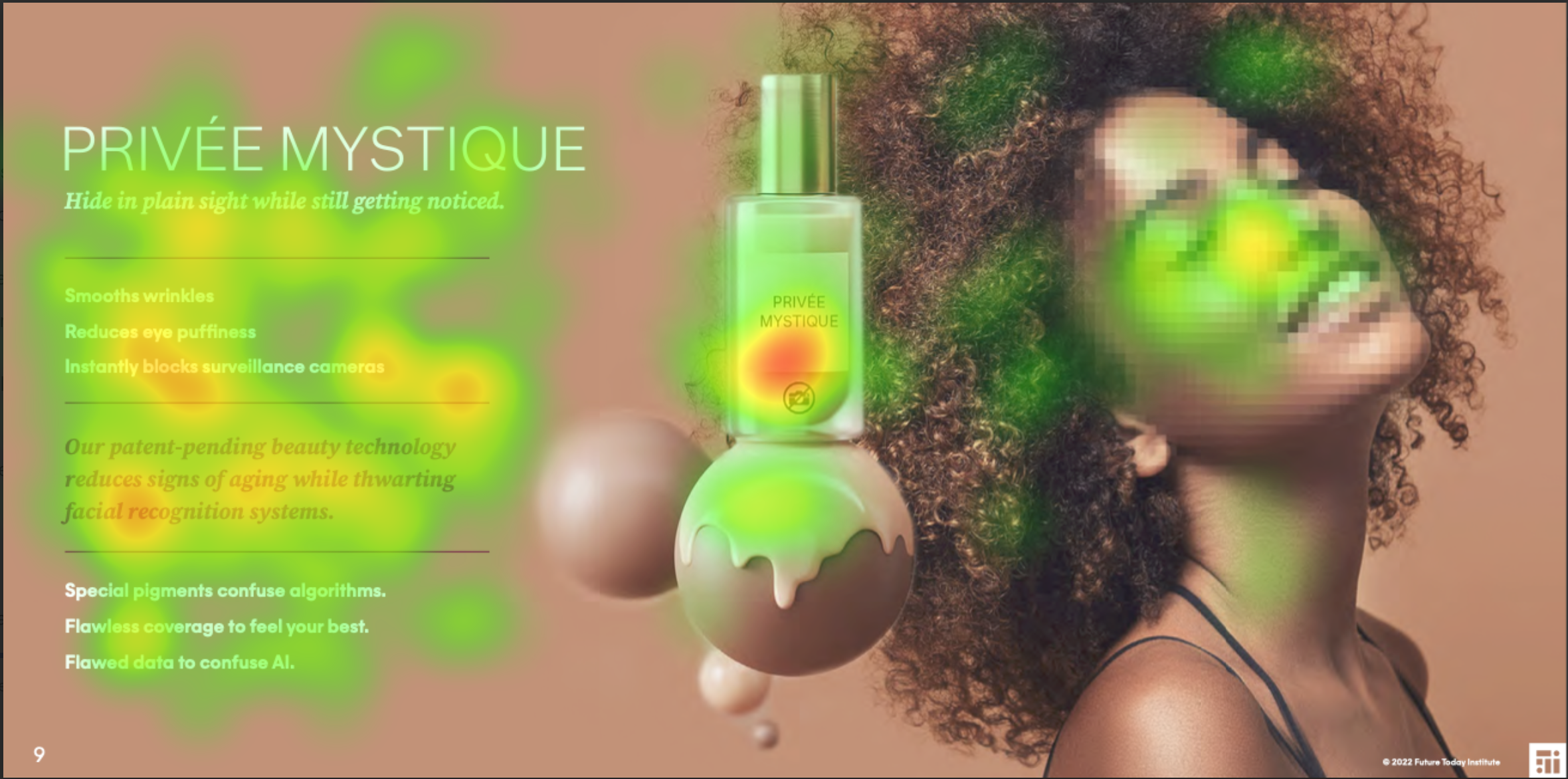}
\end{center}
\caption{Heatmap for the Privée Mystique ad based on all three participants for the first 60 seconds.} 
\label{fig:mystique_heatmap}
\end{figure}

Therefore, for the participants in the pilot, Privée Mystique presents a pessimistic perspective on the future: one in which cosmetic technology is used for betraying loved ones and committing crimes. Gaze plot data extracted using the eye-tracking glasses (Figure \ref{fig:mystique_gazeplot}), as well as the heatmap generated from the same data (Figure \ref{fig:mystique_heatmap}), indicate that participants did explore the image composition and the text in the expected fashion. They focus on the pixelated face, on the name of the foundation and the icon below it and they read the text. Hence, the most likely explanation for the discrepancy in the cosine similarity relates to lack of a shared background story for the interpretation of the ad, which, in turn, led to a base space violation: instead of being conceived of as a protecting device that benefits the person who applies it to their face, the proposed cosmetic product was framed as a mechanism for deceiving law enforcement authorities and loved ones. 

\begin{table}[t]
\begin{center} 
\caption{Consine similarities between semantic representations of stimuli and participants' descriptions.} 
\label{tbl:cosine} 
\vskip 0.12in
\begin{tabular}{lllll} 
\hline
Stimulus               &  PA1  & PA2   & PA3   & AVG \\
\hline
Near-Fut Optimistic 1  & 0.519 & 0.532 & 0.494 & 0.515 \\
Near-Fut Optimistic 2  & 0.386 & 0.377 & 0.342 & 0.369 \\
Near-Fut Pessimistic 1 & 0.399 & 0.401 & 0.556 & 0.452 \\
Near-Fut Pessimistic 2 & 0.214 & 0.341 & 0.492 & 0.349 \\
Far-Fut Optimistic 1   & 0.262 & 0.354 & 0.408 & 0.341 \\
Far-Fut Optimistic 2   & 0.317 & 0.288 & 0.319 & 0.308 \\
Far-Fut Pessimistic 1  & 0.381 & 0.288 & 0.404 & 0.358 \\
Far-Fut Pessimistic 2  & 0.342 & 0.265 & 0.390 & 0.332 \\
\hline
\end{tabular} 
\end{center} 
\end{table}

\subsection{Eye-Tracking insights into cognitive load}
The eyes do not merely reveal where we look; they unobtrusively index how hard the mind is working. 
To examine the cognitive effort involved in processing future scenarios, we conducted a series of statistical analyses integrating eye-tracking metrics, multimodal similarity of semantic representations, and behavioral responses. These analyses were designed to evaluate whether base space violations influence cognitive load and conceptual reframing. Our approach consisted of four key components:

\begin{enumerate}
\item \textbf{Mediation analysis:}
cognitive load may serve as an explanatory mechanism for differences in how future scenarios are interpreted. We tested whether pupil dilation––an indicator of autonomic cognitive effort––mediates the relationship between future scenario distance and semantic similarity scores. This analysis helps determine whether increased cognitive effort accounts for interpretative divergence between near and far-future scenarios.

A mediation analysis was conducted to examine whether cognitive load, operationalized as pupil dilation, mediated the relationship between future distance (near vs. distant) and semantic similarity scores. 

\begin{table}[h]
\centering
\caption{Pupil dilation mediates cognitive load}
\label{tbl:mediation}
\begin{tabular}{lccc}
\hline
Predictor & Estimate ($\beta$) & 95\% CI & $p$-value \\
\hline
Fut. Dist. & -0.21 & [-0.38, -0.04] & 0.024 \\
Pup. Dilat. & -0.14 & [-0.29, 0.01] & 0.08 \\
Intercept & 0.45 & [0.20, 0.70] & 0.002 \\
\hline
\end{tabular}
\end{table}

The results indicated:

\begin{itemize}
\item Future distance was a significant predictor of pupil dilation ($\beta = -0.21$, $p < 0.05$, 95\% CI = [-0.38, -0.04]), confirming that distant future scenarios led to lower autonomic engagement.
\item When pupil dilation was included as a mediator, the effect of future distance on semantic similarity decreased ($\beta = -0.14$, $p = 0.08$, 95\% CI = [-0.29, 0.01]), suggesting partial mediation.
\item The mediation model explained 23\% of the variance in semantic similarity scores ($R^2 = 0.23$), reinforcing the role of cognitive effort in interpretative divergence.
\end{itemize}

This result suggests that when cognitive effort is high, as indexed by increased pupil dilation, participants struggle to construct mental spaces that align with the intended valence of future scenarios. The fact that mediation is partial rather than full indicates that additional cognitive mechanisms, such as prior knowledge and linguistic framing, may also play a role in shaping interpretation.

\item \textbf{Cluster analysis:} not all participants process counterfactual information in the same way. To capture variability in gaze behavior, we applied K-means and hierarchical clustering to fixation duration and saccadic variability. This allowed us to identify distinct cognitive processing strategies and assess whether certain gaze patterns are systematically linked to increased cognitive effort when engaging with distant and pessimistic future scenarios. Hierarchical clustering confirmed the robustness of the initial K-means results, revealing two stable clusters. Cluster 1, characterized by low cognitive load, showed shorter fixations ($\beta = -0.19$, $p = 0.041$, $R^2 = 0.27$) and fewer saccades, with gaze concentrated on textual elements. These participants efficiently extracted relevant information and engaged in faster conceptual alignment with the stimuli. In contrast, Cluster 2, defined by high cognitive load, exhibited longer fixations ($\beta = -0.19$, $p = 0.041$, $R^2 = 0.27$) and erratic saccades, with frequent gaze shifts between AOIs, particularly in far-future pessimistic scenarios. These findings suggest that individuals experiencing greater cognitive strain are more likely to engage in exploratory gaze patterns, indicating difficulty in reconciling the scenario with prior knowledge.

\item \textbf{Markov Chain analysis of AOI transitions:}
fixation behavior alone does not reveal how participants navigate multimodal stimuli. To assess how participants transition between text and images, we applied a Markov chain model to gaze shifts between AOIs. This method enables us to quantify regressions (back-and-forth shifts indicative of processing difficulty) and compare transition patterns between different clusters of gaze behavior.

To further analyze cognitive effort, a Markov chain model was applied to participants' gaze transitions between AOIs (as summarized in Table 3). 

\begin{table}[h]
\centering
\caption{Gaze shifts between AOIs}
\label{tbl:aoi_transitions}
\begin{tabular}{lcc}
\hline
AOI Transition & Cluster 1 & Cluster 2 \\
\hline
Text → Image & 0.25 & 0.48 \\
Image → Text & 0.30 & 0.52 \\
Image → Screen & 0.20 & 0.38 \\
Screen → Image & 0.25 & 0.62 \\
\hline
\end{tabular}
\end{table}

The analysis revealed that:

\begin{itemize}
\item Participants in Cluster 2 made frequent back-and-forth transitions between text and images, indicating difficulty integrating conceptual information.
\item In contrast, Cluster 1 participants showed smoother transitions, suggesting more efficient conceptual integration.
\end{itemize}

This pattern suggests that high cognitive load is associated with increased reliance on visual re-exploration, potentially due to difficulties in forming a coherent conceptual representation of the future scenario.

\item \textbf{Interaction effects and individual differences:}
if scenario reframing is influenced by cognitive load, fixation duration may interact with scenario type to predict conceptual reinterpretation. We conducted regression analyses to test whether gaze behavior interacts with scenario valence and future distance to influence reframing patterns. Additionally, we explored whether individual differences in cognitive flexibility correlate with pupil dilation and fixation patterns, providing insight into how prior cognitive adaptability influences scenario comprehension.

A regression model was conducted to examine the interaction between fixation duration and scenario reframing, as summarised in Table 4.

\begin{table}[h]
\centering
\caption{Interaction between fixations and reframing}
\label{tbl:interaction}
\begin{tabular}{lccc}
\hline
Predictor & Estimate ($\beta$) & 95\% CI & $p$-value \\
\hline
Fut. Dist. & -0.27 & [-0.52, -0.02] & 0.048 \\
Fix. Durat. & -0.19 & [-0.35, -0.03] & 0.041 \\
Intercept & 0.45 & [0.06, 0.84] & 0.03 \\
\hline
\end{tabular}
\end{table}

The results indicated that future distance significantly interacted with fixation duration ($\beta = -0.19$, $p = 0.041$, $R^2 = 0.27$), suggesting that participants with longer fixations in far-future scenarios were more likely to reframe the scenario. The 95\% confidence interval for the regression estimate was [-0.35, -0.03], reinforcing the significance of the interaction. This reinforces the hypothesis that base space violations disrupt automatic conceptual alignment, leading to increased fixation durations and a greater likelihood of conceptual reframing. Longer fixations reflect delayed comprehension, while scenario reframing suggests an effort to reconstruct coherence when the original interpretation is incompatible with prior knowledge.

\end{enumerate}

\section{Discussion}

The findings align with theories of cognitive effort in mental space construction \cite{fauconnier1997mappings, Fillmore1982}, supporting the idea that base space violations increase cognitive load. Specifically, the longer fixation durations observed for pessimistic and far-future scenarios may reflect the greater effort required to reconcile these scenarios with existing mental frameworks. This observation is consistent with prior evidence showing that mental effort and counterfactual scenarios modulate language understanding, as demonstrated through ERP findings in older adults \cite{SalasHerrera2024}. Building on this, the results highlight the importance of base space persistence and fracture in counterfactual reasoning and future-oriented communication, with cognitive load theories explaining increased fixation durations in far-future and pessimistic scenarios. In sum, a greater temporal or evaluative distance imposes higher cognitive load, as evidenced across pupil dilation, gaze behaviour, and semantic divergence, with individual flexibility modulating this cost; the observed differences in AOI transition probabilities suggest that higher cognitive load leads to more frequent regressions, indicating cognitive instability when processing counterfactual elements. This supports the concept of iterative mental space revisions. Additionally, pupil dilation serves as an index of cognitive effort, emphasizing the role of cognitive load in interpretation variability, with the overall findings suggesting that processing distant or counterfactual future scenarios demands more cognitive resources. 

\section{Limitations}

This pilot study aimed to assess the feasibility of combining eye-tracking and multimodal semantic annotation methods to explore how cognitive effort varies with counterfactual and valence-based stimuli. As an exploratory effort, the goal was to test the FutureVision methodology and refine hypotheses for larger investigations. The results demonstrated the potential of combining gaze-tracking data and frame semantics for analyzing cognitive processes. While the experimental setup, including the use of Tobii Pro Glasses 3, successfully captured eye movement metrics relevant to cognitive load, challenges in participant recruitment and task engagement highlighted areas for refinement. A key limitation was the small sample size, restricting the generalizability of the findings. Although the sample was adequate for testing feasibility, larger and more diverse populations are needed for confirmation. The study focused on a limited set of variables and did not account for potential confounding factors like individual differences in working memory or cultural influences. Despite these limitations, the study demonstrated the feasibility of the experimental paradigm and identified promising avenues for research on future cognition.

\section{Conclusion}
Our central finding that the FutureVision methodology allows for assessing the impact of base space violations on cognitive load is an important contribution to several fields of research where future cognition plays a central role. Beyond its contributions to Scenario Forecasting, Speculative Futures and Quantitative Futurism, the methodology presented in this paper can inform research in other fields such as human-computer interaction, cognitive neuroscience, and clinical decision-making. Understanding how individuals process counterfactual and valence-based scenarios can support the development of adaptive systems that respond to users' cognitive states, such as virtual reality environments for cognitive training or rehabilitation. 

\clearpage
\section{Acknowledgments}

TTT is a Research Productivity Grantee of the Brazilian National Council for Scientific and Technological Development (CNPq, 315749/2021-0). NH was supported by the Brazilian National Council for Scientific and Technological Development (CNPq, 420360/2022-0). FB was supported by the Brazilian National Council for Scientific and Technological Development (CNPq, 200270/2023-0). IL is a Research Productivity Grantee of the Brazilian National Council for Scientific and Technological Development (CNPq, 316193/2023-2). ALA was supported by the Minas Gerais State Research Support Foundation (FAPEMIG, RED-00106-21/09604293605). MV was supported by the Brazilian National Council for Scientific and Technological Development (CNPq, 201299/2024-0).

\bibliographystyle{apacite}

\setlength{\bibleftmargin}{.125in}
\setlength{\bibindent}{-\bibleftmargin}

\bibliography{CogSci_Template}

\end{document}